\documentclass{article}

\usepackage{spconf}
\usepackage{amsmath}
\usepackage{amssymb}
\usepackage[pdftex]{graphicx}
\graphicspath{{./Figures/}}
\usepackage{subfigure}
\usepackage{multirow}
\usepackage{url}

\title{Co-Regularized Deep Representations for Video Summarization}
\name{Olivier Mor\`ere{$^{*,1,2,3}$}, Hanlin Goh{$^{*,1,3}$}, Antoine Veillard{$^{*,2,3}$}, Vijay Chandrasekhar{$^{1,3}$}, Jie Lin{$^{1,3}$}\thanks{{$*$}\ \,O. Mor\`ere, H. Goh and A. Veillard contributed equally to this work.}\thanks{1.\ Institute for Infocomm Research, A*STAR, Singapore.}\thanks{2.\ Universit\'e Pierre et Marie Curie, Paris, France.}\thanks{3.\ Image \& Pervasive Access Lab, UMI CNRS 2955, Singapore.}}
\address{I2R{$^{1}$}, UPMC{$^{2}$}, IPAL{$^{3}$}}
\begin{document}
\maketitle
\begin{abstract}

Compact keyframe-based video summaries are a popular way of generating viewership on video sharing platforms.
Yet, creating relevant and compelling summaries for arbitrarily long videos with a small number of keyframes is a challenging task.
We propose a comprehensive keyframe-based summarization framework combining deep convolutional neural networks and restricted Boltzmann machines.
An original co-regularization scheme is used to discover meaningful subject-scene associations.
The resulting multimodal representations are then used to select highly-relevant keyframes.
A comprehensive user study is conducted comparing our proposed method to a variety of schemes, including the summarization currently in use by one of the most popular video sharing websites.
The results show that our method consistently outperforms the baseline schemes for any given amount of keyframes both in terms of attractiveness and informativeness.
The lead is even more significant for smaller summaries.

\end{abstract}

\section{Introduction}
\label{sec:intro}

Video sharing websites measure user engagement through click rates and viewership.
To make a novel video attractive for the audience, its video link is often presented as a thumbnail of either a single representative frame or a slideshow of several keyframes.
In this work, we explore the problem of automatically generating diverse, representative and attractive keyframe-based summaries for videos.

\begin{figure}[ht]
\centering
\includegraphics[width=\hsize]{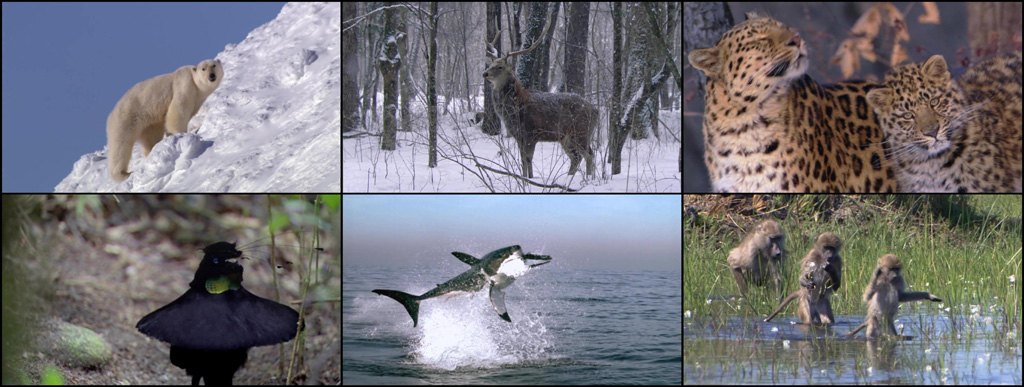}
\caption{Deep co-regularized keyframe summary. Our method extracts diverse, representative and attractive keyframes.}
\label{fig:frontsummary}
\end{figure}

Summarization-based techniques can be broadly divided into three categories: 1) keyframe-based, 2) skimming-based and 3) story-based.
In keyframe-based summarization, the video is summarized using a small number of keyframes selected based on some criterion, such as low-level features, like pixel data, motion features, optical flow and frame differences~\cite{motion1,motion2,imagediff1}, or higher-level information, like objects and faces~\cite{object1,object2}.
For this class of algorithms, clustering techniques like k-means are popular: clustering or grouping is performed based on raw RGB pixels, or a combination of low and high level features~\cite{kmeans1,kmeans2,kmeans3,kmeans4,kmeans5}.
The frames closest to the cluster centers are chosen to be part of the summary.

Skimming-based summarization is used to produce longer video summaries.
The video is divided into smaller shots using shot boundary detection algorithms and a series of shots are selected to form the summary video. Subshot selection is based on motion activity~\cite{dynamic1,dynamic2,dynamic3} and other high level features, such as person and landmark descriptors~\cite{gygli2014creating}.

Finally, in storyboard-based summarization, algorithms take into account relationships between the different subshots~\cite{storyboard1}. This enables long egocentric videos to be summarized to gain an understanding of the underlying events.

{\bf Contributions.} This work focuses on generating compact keyframe-based summarization,
with the main contributions are as follows:\\[-1.75em]
\begin{itemize}
\item A comprehensive keyframe-based summarization framework combining deep convolutional neural networks (DCNNs) and restricted Boltzmann machines (RBMs).\\[-1.75em]
\item A co-regularization scheme for restricted RBMs able to learn joint high-level subject-scene representations.\\[-1.75em]
\item A comprehensive user study comparing our method against various schemes including the algorithm in use by the video sharing website \emph{Dailymotion}.\\[-1.75em]
\end{itemize}

\section{Co-Regularized Deep Representations}
\label{sec:summarization}

A good keyframe-based summary should consist of easily recognizable subjects in context-setting scenes.
To achieve this, we generate frame-level descriptions by exploiting deep convolutional architectures to recognize subjects and scenes.
Compact representations are then computed with a novel co-regularization unsupervised learning scheme to exhibit the high-level associations between subjects and scenes.
Keyframes are subsequently generated from these compact representations.

\subsection{Deep Convolutional Neural Networks}
\label{sec:cnn}

Deep convolutional neural networks (DCNNs) have recently been used to obtain astonishing performances in both image classification \cite{Krizhevsky2012b,Razavian2014} and image retrieval \cite{Cvpr2015} tasks.
For every frame sampled from the video at regular intervals, DCNN descriptors are extracted using the open source Caffe framework \cite{Jia2014} along with two pre-trained networks: VGG-ILSVRC-2014-D \cite{Simonyan2014} and Places-CNN \cite{Zhou2014}.

VGG-ILSVRC-2014-D is the best performing single network from the VGG team during the ILSVRC 2014 image classification and localization challenge using the ImageNet \cite{Deng2009} dataset.
This 138 million parameters network is made of 16 layers: 13 convolutional layers followed by 3 fully-connected layers.
It detects 1000 mostly subject-centric categories (e.g. animals, objects, plants, etc\ldots).

Places-CNN is a 60 million parameters network following the AlexNet \cite{Krizhevsky2012b} structure: a total of 8 layers: 5 convolutional layers followed by 3 fully-connected layers.
It is trained on the Places 205 dataset, a scene-centric image dataset featuring 205 categories including indoors and outdoors sceneries.

For both DCNNs, descriptors are extracted from the last layer before the softmax operation, having a dimensionality of 1000 and 205 for VGG-ILSVRC-2014-D and Places-CNN, respectively.

\subsection{Co-Regularized Restricted Boltzmann Machines}
\label{sec:rbm}

To create the video summaries from the DCNN descriptions of subjects $\mathbf{x}_{o}$ and scenes $\mathbf{x}_{p}$, we introduce a pair of concurrently trained restricted Boltzmann machines (RBMs) to learn their projections ($\mathbf{z}_{o}$ and $\mathbf{z}_{p}$) to $K$ units each, where $K$ is the desired number of keyframes. An RBM is a bipartite network with a projection matrix $\mathbf{W}$ that maps between its input and output units. RBMs are trained through gradient descent on the approximate maximum likelihood objective, based on network states drawn from Gibbs sampling~\cite{hinton-02,hinton-06}.

\begin{figure}[ht]
\centering
\subfigure[Training co-regularized RBMs.]{
\includegraphics[width=\hsize]{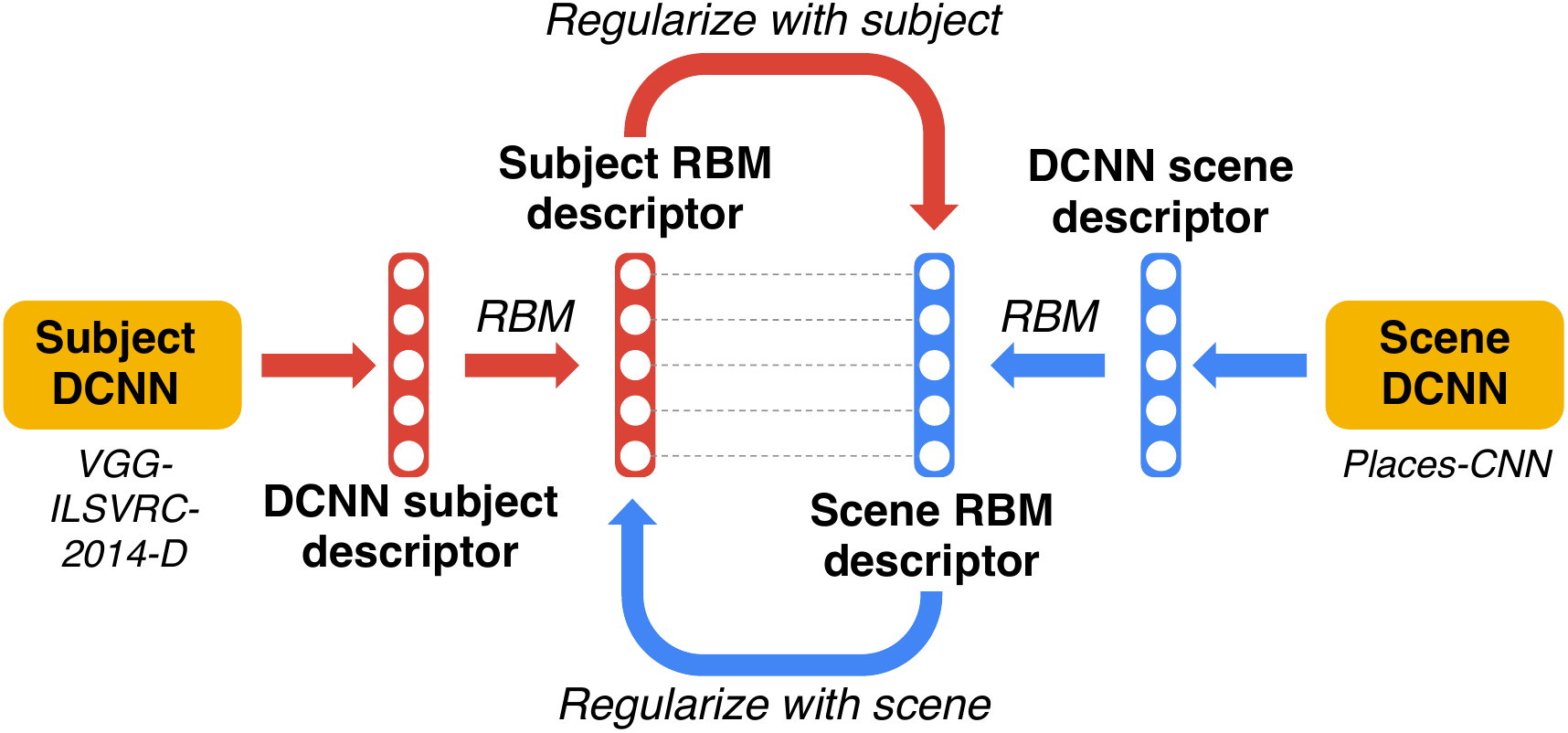}
\label{fig:crrbmtrain}
}
\subfigure[Generating a frame descriptor from co-regularized RBMs.]{
\includegraphics[width=\hsize]{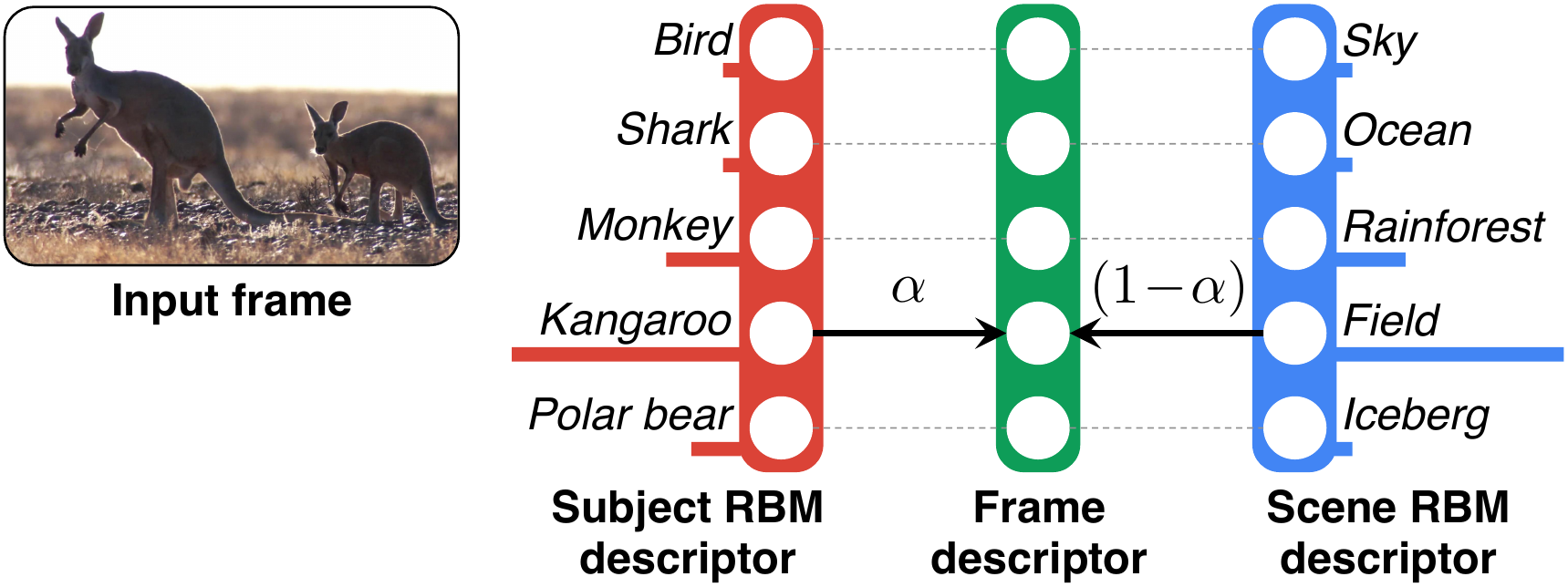}
\label{fig:framescore}
}
\caption{A pair of co-regularized RBMs -- one representing subjects and another representing scenes -- are learned concurrently.
\subref{fig:crrbmtrain} During training, an subject unit is regularized by its corresponding scene unit and vice versa.
\subref{fig:framescore} The frame descriptor is a linear combination of the two co-regularizated RBM descriptors forming relevant subject-scene associations.}
\label{fig:crrbm}
\end{figure}

In this work, we introduce co-regularization for RBMs.
The object RBM is regularized by place representations and in turn regularizes the training of the place RBM (Figure~\ref{fig:crrbmtrain}).
Given randomly sampled minibatches of subject and scene DCNN descriptors $\{\mathcal{X}_{o}^{i},\mathcal{X}_{p}^{i}\}_i$, we introduce co-regularization cross entropy penalties to the RBM objective functions:
\begin{equation}\label{eq:objectloss}
\underset{\mathbf{W}\!_{o}}{\arg\min}\!-\!\!\sum_{i}\!\sum_{\mathbf{z}_{o}^{i}\in\mathcal{Z}^{i}_{o}}\!\!\bigg(\!\log\mathbb{P}(\mathbf{x}^{i}_{o},\mathbf{z}_{o}^{i})\!-\!\lambda_{o}\!\sum_{k}\log\mathbb{P}(\hat{z}_{p,k}^{i}\!\mid\! \hat{z}_{o,k}^{i})\!\bigg),
\end{equation}
\begin{equation}\label{eq:placeloss}
\underset{\mathbf{W}\!_{p}}{\arg\min}\!-\!\!\sum_{i}\!\sum_{\mathbf{z}_{p}^{i}\in\mathcal{Z}^{i}_{p}}\!\!\bigg(\!\log\mathbb{P}(\mathbf{x}^{i}_{p},\mathbf{z}_{p}^{i})\!-\!\lambda_{p}\!\sum_{k}\log\mathbb{P}(\hat{z}_{o,k}^{i}\!\mid\! \hat{z}_{p,k}^{i})\!\bigg),
\end{equation}
where  $\{\mathcal{Z}_{o}^{i},\mathcal{Z}_{p}^{i}\}_i$ are the RBM projections of $\{\mathcal{X}_{o}^{i},\mathcal{X}_{p}^{i}\}_i$, $\{\lambda_{o},\lambda_{p}\}$ are the regularization constants, and $\{\hat{z}_{o,k}^{i}, \hat{z}_{p,k}^{i}\}_i$ refer to unit $k$ in the distribution-sparsified representations of the minibatch~\cite{goh-12}. Sparsity across units helps avoid co-adaptation between the units and improves representational diversity across instances of frames.
The co-regularization terms serves the purpose of binding a subject and the scene in which it occurs to the same unit position.

The frame descriptor is a linear combination of the two RBM descriptors (Figure~\ref{fig:framescore}). The final set of keyframe timings $t_k, k \in [1..K]$ is the ordered set of $K$ timings that gives the maximum response for each unit of the frame descriptor:
\begin{equation}
\underset{t}{\arg\max}\ \ \alpha\,z_{o,k}^{t}+(1-\alpha){z}_{p,k}^{t},
\label{eq:alpha}
\end{equation}
where $\alpha \in [0,1]$ is a balance hyperparameter that causes the summary to be more subject-centric or scene-centric.

This proposed co-regularization method is not specific to subjects or scenes, and is generalizable to other concepts or modalities, such as faces or activities.

\section{Video Summarization}
\label{sec:results}
Using our method, we summarized all 11 episodes from the BBC educational TV series \emph{Planet Earth}\footnote{\url{http://www.bbc.co.uk/programmes/b006mywy}}. Each episode is approximately 50 minutes long. A sample of our results is shown in Figure~\ref{fig:ourmethod}.

\subsection{Model Visualization}
\label{sec:visualization}

\subsubsection{Balancing Subject- and Scene-Centricity}

\begin{figure}[ht]
\centering
\includegraphics[width=\hsize]{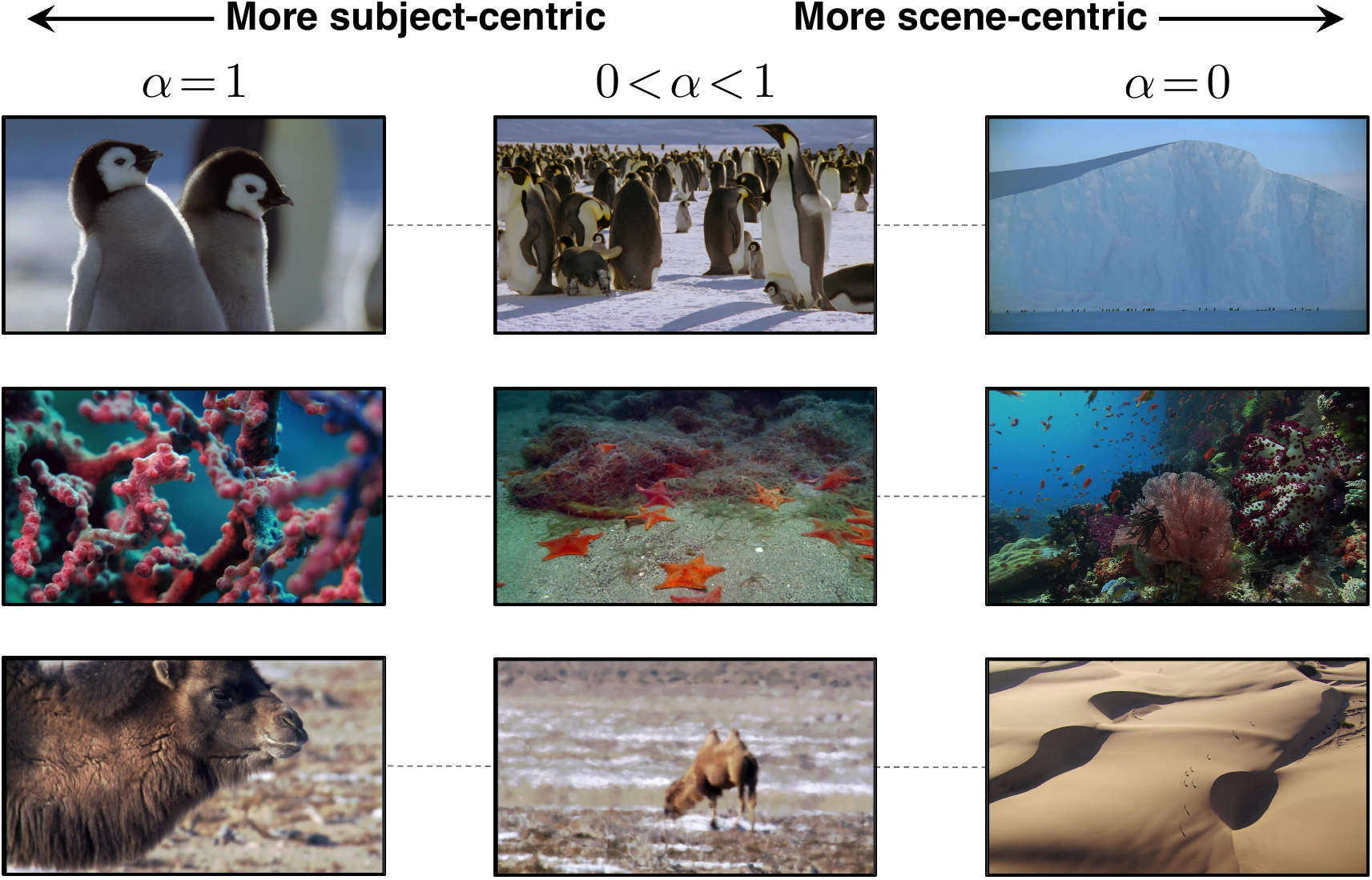}
\caption{Actual keyframes selected by varying $\alpha$. Our model can be tuned to select keyframes that are more subject-centric (left), scene-centric (right) or a balance of both (middle).}
\label{fig:subjectscene}
\end{figure}

As shown in Figure~\ref{fig:subjectscene}, bias towards subject or scenes can be adjusted by tuning the $\alpha$ parameter from Equation~\ref{eq:alpha}.
This flexibility allows for interesting functionalities such as customising content based on user profiling or explicit queries. 
The choice of $\alpha$ value can also be made independently for each unit in order to generate the most visually attractive keyframe, for example based on vibrancy.
In practice, setting the default value to $\alpha=0.5$ (as used in this empirical study) seems to produce satisfactory results.

\subsubsection{Visualisation of Co-Regularized RBM Units}

\begin{figure}[ht]
\centering
\includegraphics[width=\hsize]{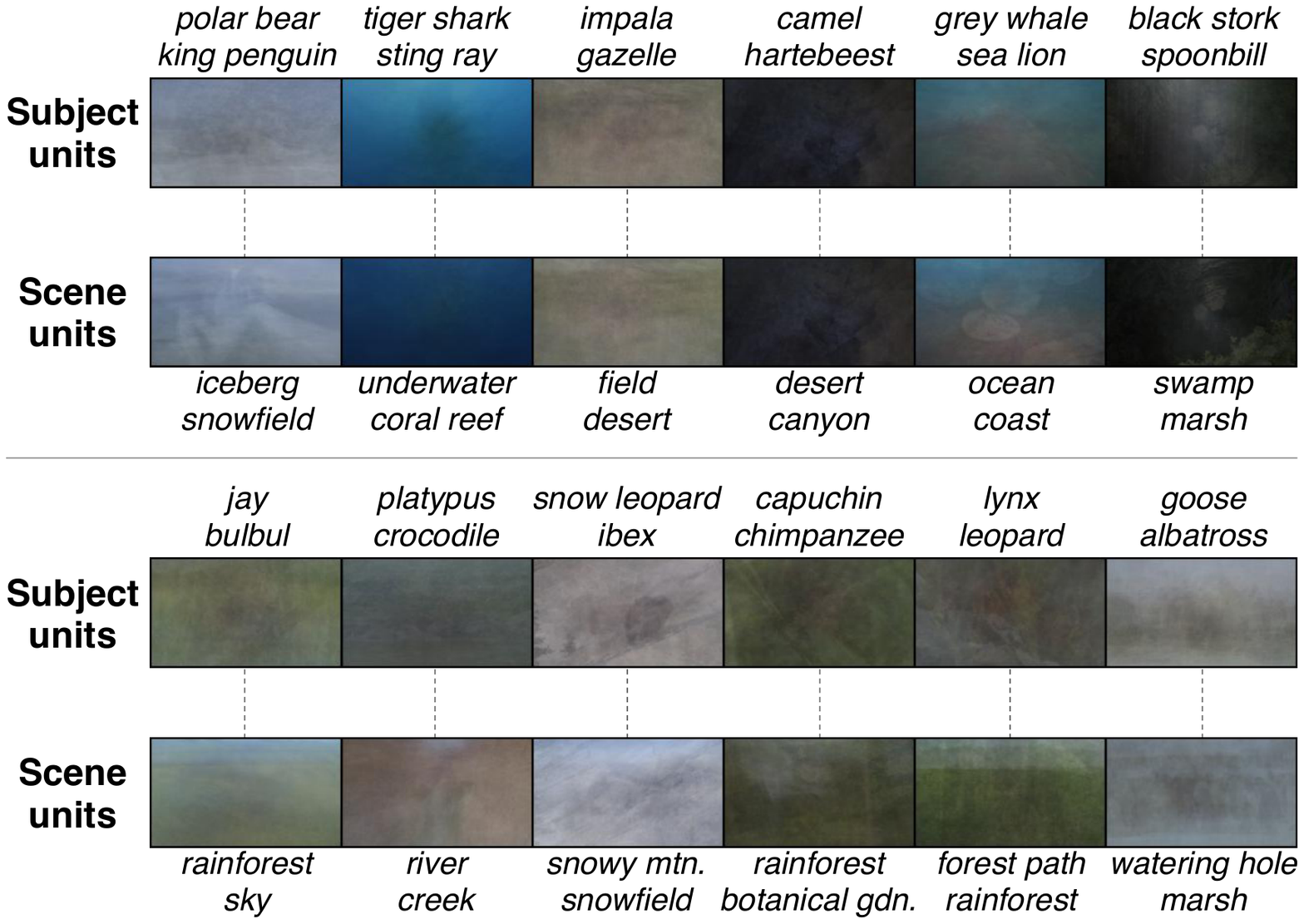}
\caption{Visualization of the units for a $K=12$ model.
The visual representations of subject-scene pairs are well correlated.
The categories of the two models are associated in a sensible way and correspond well with the visual representations.
}
\label{fig:units}
\end{figure}

Although neural networks tend to be thought of as black boxes, visualization is often useful to dechipher what has been learned~\cite{zeiler14}.
To better understand our co-regularized model, we analysed the responses of each unit across the dataset.
For this analysis, we trained a single $K=12$ model across all 11 episodes.
For each of the 24 RBM units, the top 100 frames that most strongly activate each unit were aggregated via a weighted average.
The resulting graphical representation of each unit is shown on Figure~\ref{fig:units}.
We observe that the visual appearances of frames corresponding to a subject-scene pair of units are consistently similar.
There is also diversity across the units within an RBM.

The top 2 categories of each unit identified from the weight matrices are also shown in Figure~\ref{fig:units}.
We notice that the correlation with the visual representation is strong and the subject-scene association is sensibly learned.
We can also observe an interesting effect of co-regularization, where associations can be made between subjects (e.g. \emph{polar bear} and \emph{king penguin}) that occur in the same scene (\emph{iceberg}) but never within the same frame.

\subsection{User Engagement Study}
\label{sec:userstudy}

\subsubsection{Evaluation Framework}
\label{sec:evaluationframework}

\begin{figure*}[ht]
\centering
\subfigure[Our method]{\includegraphics[width = 2\columnwidth]{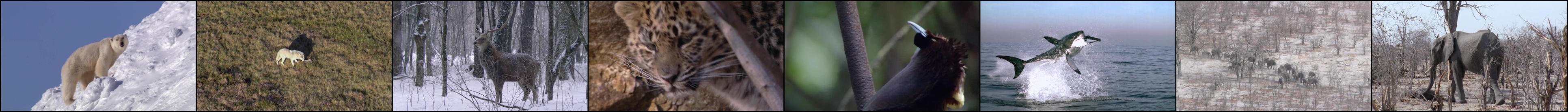}\label{fig:ourmethod}}
\subfigure[Uniform sampling]{\includegraphics[width = 2\columnwidth]{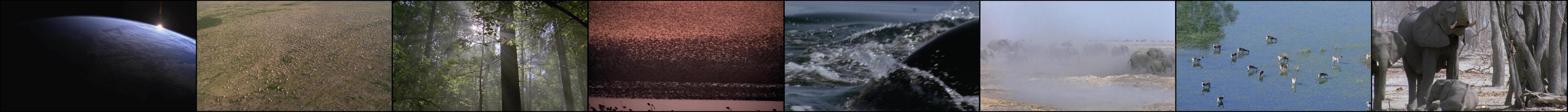}}
\subfigure[k-means]{\includegraphics[width = 2\columnwidth]{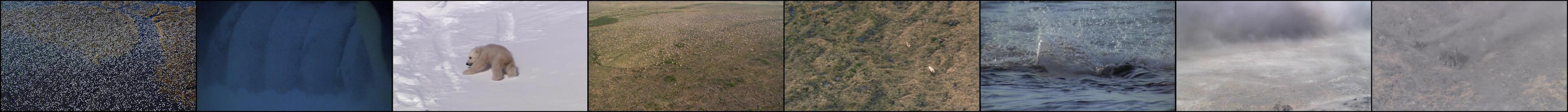}}
\subfigure[Dailymotion]{\includegraphics[width = 2\columnwidth]{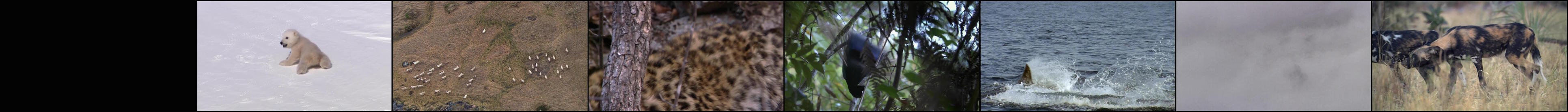}}
\caption{Eight keyframes summaries for episode 1 from the TV series \emph{Planet Earth}.}
\label{fig:4schemes}
\end{figure*}

Our method is compared against three other keyframe-based summarisation schemes: naive uniform sampling, k-means clustering and the method currently in use by the video sharing website \emph{Dailymotion}\footnote{\url{http://www.dailymotion.com/}}.
Each summary is presented as a timeline of keyframes as shown on Figure~\ref{fig:4schemes}.

Uniform sampling takes $k$ keyframes with evenly spaced timestamps: $t_i = \frac{d}{k}\left( \frac{1}{2} + i \right),i \in [1..K]$ where $d$ is the total duration of the video.
The k-means clustering scheme uses frames sampled at the same frequency as for our method (1 {\it fps}) and down-sized to $32 \times 32$ RGB pixels.
Lloyd's algorithm~\cite{lloyd1982least} is used to separate the data into $K$ clusters.
100 runs with different centroid seeds are performed to mitigate the effects of local minima.
For each cluster, the frame closest to its centroid is selected as keyframe.
\emph{Dailymotion} proposes an 8-keyframes video summary (excluding title frame) which was used as a blackbox scheme to compare our method against.
The evaluation videos were uploaded on the website and the proposed summary keyframes were then handpicked from the original footages.

The study was performed by showing pairs of summaries -- our method against one of the three baseline schemes -- to eight different testers who have not previously seen the videos.
For each pair, they are asked to answer the two following questions:
\begin{itemize}
\item Q1: \emph{Which video would you rather watch?\\(attractiveness)}\\[-1.75em]
\item Q2: \emph{Which summary was more informative?\\(informativeness)}
\end{itemize}

Using all the 11 \emph{Planet Earth} episodes, summaries were generated for different amount of keyframes $K=4,6,8$, except for \emph{Dailymotion} which imposes $K=8$ by default.
In total, $8 \times 11 \times 2 \times 3 \times + 8 \times 11 = 616$ answers were collected for each question.

Uniform sampling appears as a natural choice for the wildlife documentaries used during this study given to the slow pace of the action and high visual appeal of the average frame.
K-means is expected to be able to capture the diversity of the scenes well whereas it may not perform as well with respect to subjects.

\subsubsection{Results and Discussion}
\label{sec:resultsanddiscussion}

Table~\ref{tab:results} aggregates the answers from the testers.
Overall, our method was systematically found more attractive (75\% to 97.73\% of the time) and more informative (76.14\% to 94.32\%).
Perceived attractiveness and informativeness are strongly correlated.
Against \emph{Dailymotion}'s algorithm, our method scores favourably more than three times out of four representing a marked improvement over the scheme currently used by the service.

\begin{table}[htb]
\caption{How often our method is preferred over each of the three schemes (percentage) for different $K$.}
\label{tab:results}
\centering
\small
\scalebox{0.99}{
\hspace{-0.07in}
\vspace{-0.25em}
\begin{tabular}{|c|ccc|ccc|c|}
\hline
& \multicolumn{3}{|c|}{uniform} & \multicolumn{3}{|c|}{k-means} & daily. \\
\hline
$K$ & 4 & 6 & 8 & 4 & 6 & 8 & 8 \\
\hline
\hline
Q1 & 79.55 & 82.95 & 76.14 & 97.73 & 82.95 & 75.00 & 77.27\\
\hline
Q2 & 78.41 & 80.68 & 81.82 & 94.32 & 80.68 & 76.14 & 78.41\\
\hline
\end{tabular}
}
\end{table}

For varying amounts $K$ of keyframes, the improvement is rather consistent against uniform sampling whereas agains k-means, the improvement is more pronounced when $K$ is smaller.
This is an indication that our overall good method is particularly well-suited for compact summaries.
\vspace{-1em}
\section{Conclusions}
\label{sec:conclusions}

Building upon recent advances in deep learning and image recognition, we proposed a comprehensive keyframe-based summarization framework combining DCNNs and RBMs.
Through a comprehensive empirical study, we showed that our method is able to out perform a number of existing schemes.
In addition, our novel co-regularization scheme, which discovered meaningful subject-scene associations is generalizable to other concepts and modalities.

Beyond the selection of quality keyframes, our contribution represents a strong step towards the Holy Grail of text-based video summaries by introducing highly interpretable semantic representations.\\

\par{{\centering
\vspace{-0.25em}
{\bf\uppercase{Acknowledgements}\\[0.5em]}
}}
\noindent We gratefully acknowledge the support of NVIDIA Corporation with the donation of the GPU used for this research.

\bibliographystyle{IEEEbib}
\bibliography{refs}
\end{document}